\documentclass[conference]{IEEEtran}
\IEEEoverridecommandlockouts
\usepackage{cite}
\usepackage{amsmath,amssymb,amsfonts}
\usepackage{algorithmic}
\usepackage{graphicx}
\usepackage{textcomp}
\usepackage{xcolor}
\def\BibTeX{{\rm B\kern-.05em{\sc i\kern-.025em b}\kern-.08em
    T\kern-.1667em\lower.7ex\hbox{E}\kern-.125emX}}
\begin{document}

\title{Fast Person Detection Using YOLOX With AI Accelerator For Train Station Safety

}

\author{
\IEEEauthorblockN{
Mas Nurul Achmadiah\IEEEauthorrefmark{1}\IEEEauthorrefmark{4}\IEEEauthorrefmark{5},
Novendra Setyawan\IEEEauthorrefmark{1}\IEEEauthorrefmark{3},
Achmad Arif Bryantono\IEEEauthorrefmark{6},
Chi-Chia Sun\IEEEauthorrefmark{2}\IEEEauthorrefmark{2},
Wen-Kai Kuo\IEEEauthorrefmark{1}
}
\IEEEauthorblockA{\IEEEauthorrefmark{1}
Department of Electro-Optics, National Formosa University, Taiwan}
\IEEEauthorblockA{\IEEEauthorrefmark{2}
Department of Electrical Engineering, National Taipei University, Taiwan}
\IEEEauthorblockA{\IEEEauthorrefmark{3}
Department of Electrical Engineering, University of Muhammadiyah Malang, Indonesia}
\IEEEauthorblockA{\IEEEauthorrefmark{4}
Department of Electronics Engineering, State Polytechnic of Malang, Indonesia}
\IEEEauthorblockA{\IEEEauthorrefmark{5}
Smart Manufacturing and Intelligent Machinery Research Center, National Formosa University, Taiwan}
\IEEEauthorblockA{\IEEEauthorrefmark{6}
Department of Electronics Engineering, National Formosa University, Taiwan}
\IEEEauthorblockA{Email: chichiasun@gm.ntpu.edu.tw}
}

\maketitle

\begin{abstract}
Recently, Image processing has advanced Faster and applied in many fields, including health, industry, and transportation. In the transportation sector, object detection is widely used to improve security, for example, in traffic security and passenger crossings at train stations.
Some accidents occur in the train crossing area at the station, like passengers uncarefully when passing through the yellow line. So further security needs to be developed. Additional technology is required to reduce the number of accidents. 
This paper focuses on passenger detection applications at train stations using YOLOX and  Edge AI Accelerator hardware. the performance of the AI accelerator will be compared with Jetson Orin Nano.
The experimental results show that the Hailo-8 AI hardware accelerator has higher accuracy than Jetson Orin Nano (improvement of over 12\%) and has lower latency than Jetson Orin Nano (reduced 20 ms).
\end{abstract}

\begin{IEEEkeywords}
Fast Person Detection, YOLOX, Train Station Safety, Object detection, Classification
\end{IEEEkeywords}

\section{Introduction}
\IEEEPARstart{N}{owadays}, on the train platform, many passengers still need to pay attention when crossing the yellow safety line, and it's even more dangerous when the train almost arrives. In developing countries, train accidents are increasing day by day\cite{ref17}. There needs to be more than the existing security to protect the crowded passengers during rush hour. This proves that manual guarding has not been able to reduce accidents at train stations. Hence, an automatic passenger detection security system is needed. This system must be able to provide early warning when passengers on platforms crossing the Yellow line when a train arrives.

Currently, object detection is one of the most significant and popular topics in computer vision. Identifying items in an image and using bounding boxes to locate them are the two main objectives of object detection. Object detection technology has advanced significantly in response to the demands of the monitoring industry and the advancement of deep learning. It is extensively utilized in security, traffic control, and intelligent monitoring.\cite{ref18}. Object detection begins with the Convolutional Neural Networks (CNNs) method. This method is capable of detecting several types of objects. \textbf {However}, the detection of people in crowded environments still needs to be improved. Especially in crowded locations and need a fast process detection. 

For a variety of reasons, detecting persons are harder than a lot of other object detection. First of all, people are multifaceted and behaviour things, and it is difficult to establish a single model that all their behaviour. A powerful person detection system should be able to identify persons even when their limbs are positioned differently from one another. Second, there is a great deal of intraclass variance in the persons class due to individuals's diverse wardrobe choices (skirts, slacks, etc.), which would hinder the effectiveness of color or fine-scale edge-based approaches.\cite{ref19}

There are two types of deep learning-based object detection algorithms: models based on regression and models based on regional candidate boxes. Detectors for target detection fall into two main categories: one-stage models \cite{ref1,ref2,ref3} and two-stage models \cite{ref4,ref5,ref6}. The benefits of the single-stage type are its straightforward construction, quick operation, and balance between accuracy and efficiency.
One-stage models, like YOLOV4 and YOLOX, integrate regression and classification tasks. \cite{ref7}. Accuracy is sacrificed for processing speed in two-stage models, including Faster R-CNN \cite{ref8}, Mask R-CNN \cite{ref9}, and Mask Refined R-CNN \cite{ref10}. While some multi-object detection models, like COCO \cite{ref11} and PASCAL VOC \cite{ref12}, can achieve great performance, they are not appropriate for crowd detection, leading to an excessive number of missed detection results. The YOLO series is now the best model for single-stage detectors. YOLOX is the most versatile and effective member of the YOLO family, capable of handling a wide range of inspection tasks. YOLOX, the most recent model in the YOLO family, employs a Decoupled Head, which, in contrast to the YOLO model, separates regression and classification. 

In terms of hardware, the most widely used object detection processing currently is GPU. \textbf {However}, security warnings at train stations require high-speed processing. GPU has large power consumption and high latency. In our project, we need to detect people at train stations quickly and accurately. In every test, it has been found that the GPU operates more quickly than the CPU. \cite{ref20}. Some AI Accelerator modul also has low latency, high throughput, and low power includes AI Accelerator. In this paper we propose \textbf {A Fast person detection method using YOLOX and run in Hailo-8 AI Accelerator} so that it can provide the best solutions for safety and automation on railway lines.

\section{Method}

\subsection{Dataset}
One popular dataset for evaluating object detection and segmentation models is Microsoft Common Objects in Context (MCOCO). It is made up of real photographs of intricate settings with several objects. Compared to the well-known ImageNet dataset, MCOCO has fewer categories but significantly more images in each of those categories. MCOCO provides many items per image, accurate 2D localization of objects (bounding boxes), and non-iconic views (e.g., objects can be occluded) to solve the shortcomings of earlier datasets. \cite{ref13}. In this project, the data set used is the COCO data set This has about 140,000 pictures in it. More than 80 categories are included in the goal classification, and the majority of these categories contain more than 5,000 object instances. \cite{ref21}. Therefore, the COCO dataset is used to conduct experiments to classify the detected objects. 

\subsection{YOLOX Algorithm}
Extending YOLOv3 and YOLOv5, YOLOX \cite{ref22} is a new YOLO series algorithm that substantially enhances detection performance. The main components of YOLOX are the Head prediction module, the Neck feature pyramid, and the Backbone network. Of them, three useful feature layers—Focus, Cross Stage Partial Network (CSPnet) \cite{ref15}, and Spatial Pyramid Pooling (SPP)—are derived from input images. Cross Stage Partial (CSP) Darknet, a YOLOX backbone feature extraction network, powers the backbone network. CSP-PAFPN (PAFPN, Path Aggregation Feature Pyramid Networks) fully utilizes the three effective feature layers that the backbone network produces to build the feature pyramid by use of bottom-up and top-down feature fusion. Using the Anchor frame method based on Anchor free, which eliminates the preset priority frame and predicts the edges of the target directly to reduce computational redundancy, the prediction module improves the detection head to Decoupled Head, which increases the training convergence speed and detection accuracy of the model.

In this work, the most recent YOLOX model is applied to accomplish quick detection by deftly merging multiple good target recognition techniques. YOLOX is a traditional single-stage detection network. Unlike two-stage methods (like Faster R-CNN), the YOLO algorithm converts the detection problem into a regression problem from which bounding boxes and object confidence are immediately available. The four variations of the YOLOX model—YOLOXs, YOLOXm, YOLOXl, and YOLOXx—are obtained by varying the depth and width of the backbone network.

The input, backbone, neck, and prediction components of YOLOX architecture are its four basic components. Among these are the convolution layer (Conv), the leaky ReLU activating function, and the batch normalization layer (BN) of the basic module CBL. Multi-scale fusion benefits especially from spatial pyramid pooling (SPP) \cite{ref16}; more network modules were eventually included. Two CBL modules and an Add layer make up the residual structure, Resuit; the Add layer directly implements stacked tensors, while the Concat layer collects tensors with extended tensor dimensions.

\begin{figure}[h!]
\centering
\includegraphics[width=0.5\textwidth]{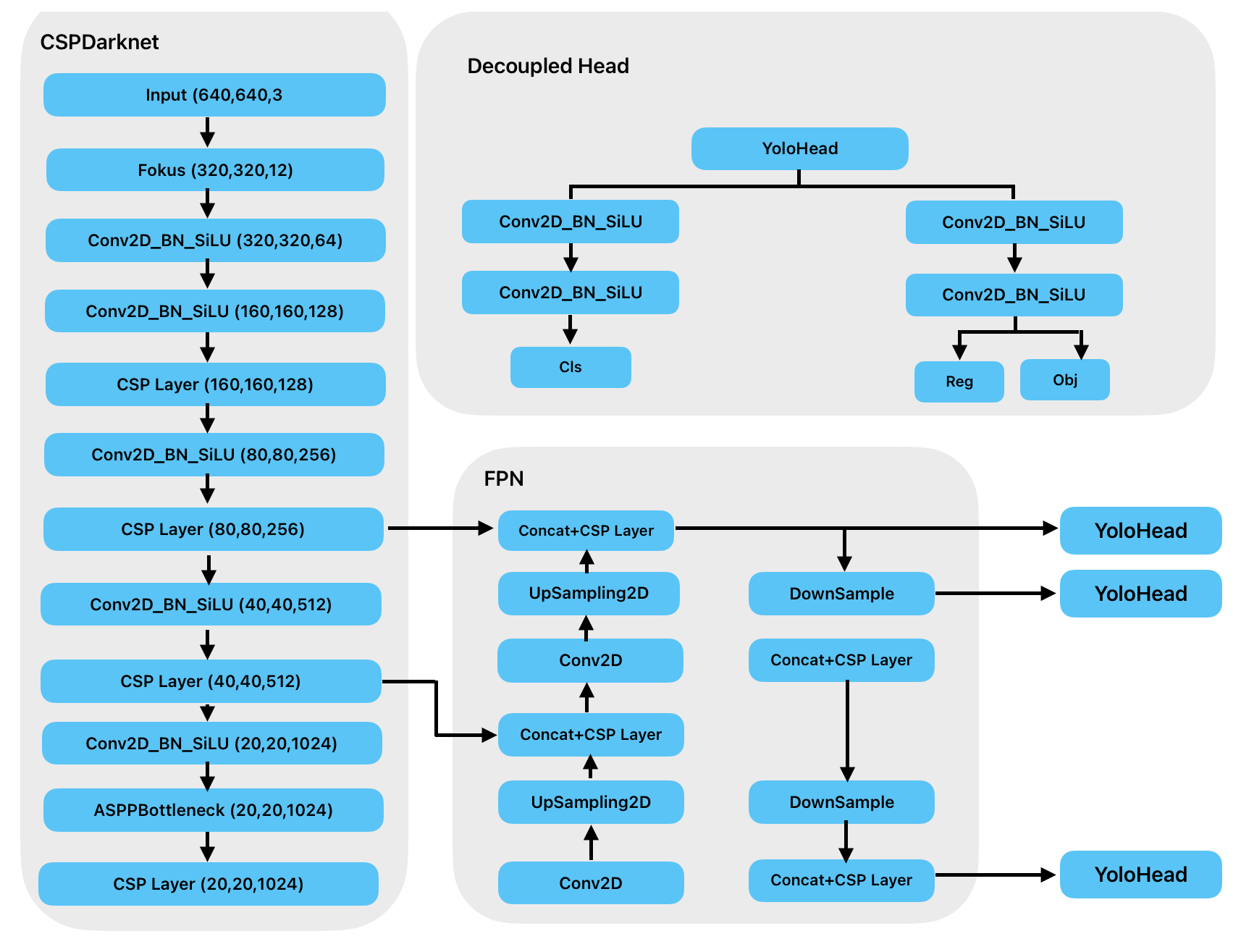}
\caption{YOLOX Structure}
\label{fig:yolox}
\end{figure}

View the three YOLOX model components at \ref{fig:yolox}. YoloHead is a network for improved feature extraction, much like the backbone network. By use of several channel modifications, the backbone feature extraction network extracts feature information from feature layers with various geometries using the CSPDarknet network. By means of the feature pyramid network, the enhanced feature extraction network facilitates feature fusion. The object kind included in the feature points, the regression parameter of the feature points if the feature points contain objects, and the regression parameter of the feature points otherwise are the three prediction results that the detection head—which consists of the classifier and regressor parts—generates.

The three enhanced effective feature layers will be obtained and fed to the detection head to determine whether the feature points have objects matching to them after the input image has undergone feature extraction in CSPDarknet and three feature layers in FPN to combine feature information at different scales for feature fusion.

\subsection{Model Build in Hailo-8\textsuperscript{TM} AI Accelerator}

Hailo, a leading AI chip maker, has redefined the edge computing landscape with its innovative AI processors and accelerators. At the heart of this revolution lies Hailo- $8^{TM}$,
an innovative AI processor for edge devices. Designed to accelerate deep learning applications on edge devices, Hailo products feature the perfect blend of high performance and power efficiency. Edge AI, which is the focal point of Hailo’s innovation, represents a significant advancement in the way data is processed and analyzed. By allowing devices to run advanced deep learning algorithms locally, Hailo processors minimize latency, which is an important aspect for applications that require real-time analysis, such as video analysis and computer vision. 

Additionally, Hailo AI processors, especially Hailo-8TM AI, are the product of rethinking traditional computer architecture. They boast unprecedented AI performance, with features that support operations at up to 26 tera per second. This extraordinary computing capability, combined with minimal power consumption, positions Hailo as a breakthrough in delivering cost-effective edge AI hardware solutions. \cite{ref22} 

Pre-trained models for high-performance deep-learning applications are available at the Hailo Model Zoo. One may measure the accuracy on the Hailo-8 device, the quantized accuracy with the Hailo Emulator, and the complete precise accuracy of each model with the Hailo Model Zoo. At last, you can produce the binary file in Hailo Executable Format (HEF) to accelerate development and produce high-quality apps accelerated by Hailo-8. The Hailo Model Zoo also offers retraining instructions for models learned on internal datasets and on custom datasets for certain use cases. Versions Hailo offers pre-compiled HEF (Hailo Executable Format) binary files for Hailo devices to run as well as several pre-trained models in ONNX / TF formats.

\begin{figure}[h!]
\centering
\includegraphics[width=0.25\textwidth]{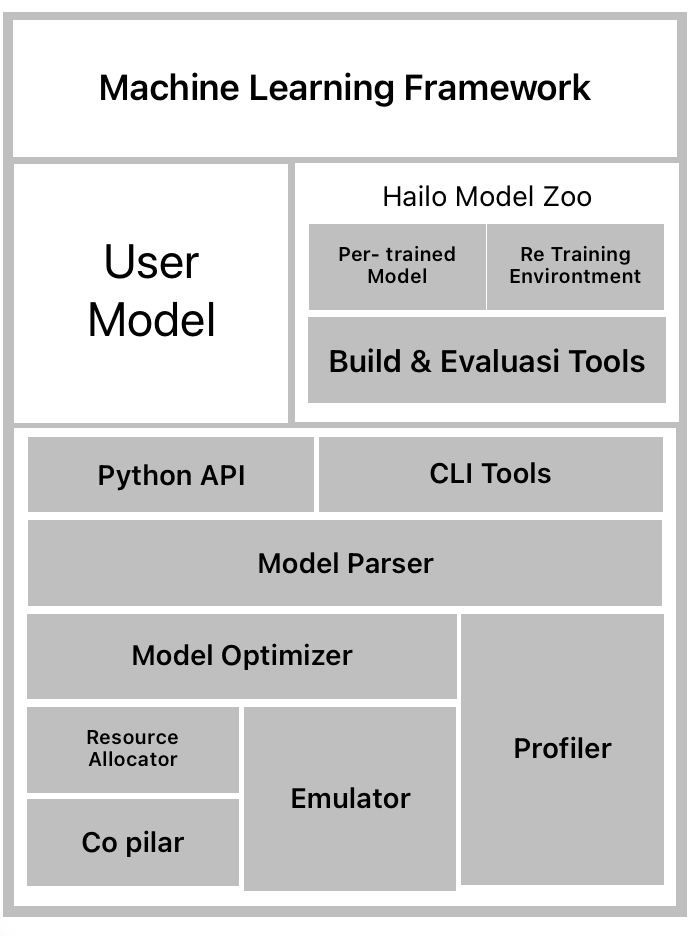}
\caption{Model Build Environment}
\label{fig:mlf}
\end{figure}

The models are divided into:
\begin{itemize}
\item Public models - which were trained on publicly available datasets.
\item HAILO MODELS which were trained in-house for specific use cases on internal datasets.
Each Hailo Model is accompanied by retraining instructions.
\end{itemize}

\subsection{Run-Time Architecture}
To use an AI processor for edge devices, we need to create a binary file that is optimized for the Hailo product. We will actually run the Python script, convert the AI model (ONNX file) learned with the deep learning development framework to Hailo, and get high performance. This project uses "Hailo AI Software Suite (2024-01)" provided by Hailo with Docker. Hailo AI Software Suite contains Hailo's main development tools in one package, making it convenient to create a development environment just by installing it. It can be downloaded for free by registering on the Developer Zone of Hailo's website.

\begin{figure}[h!]
\centering
\includegraphics[width=0.25\textwidth]{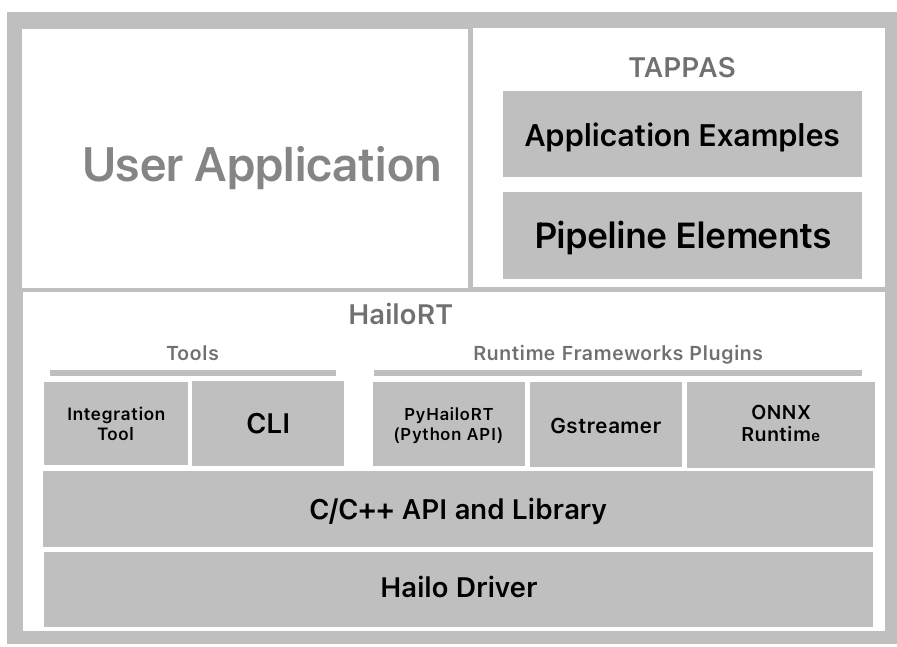}
\caption{Run-Time Architecture}
\label{fig:RE}
\end{figure}

Figure \ref{fig:mlf} explains the flow of how to create a HEF file (Hailo Executable Binary File) from an ONNX file in the Model Build Environment on the left. First copy the HEF file to the inference engine (Runtime Environment), then run inference using the HailoRT API. See Figure \ref{fig:RE}

\subsection{Fast Person Detection Algorithm }

\begin{figure}[h!]
\centering
\includegraphics[width=0.25\textwidth]{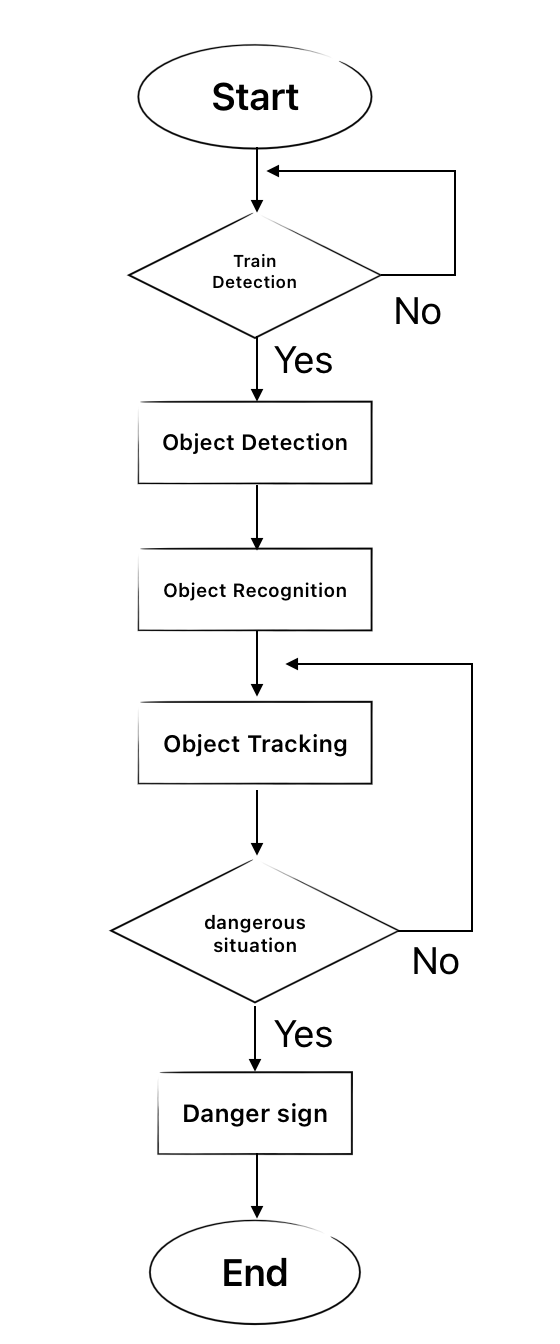}
\caption{Detection process}
\label{fig:DP}
\end{figure}

Two primary processes may be distinguished in the object detection process: training and object detection, as illustrated in figure \ref{fig:DP}. We verify the train states in the present monitoring region in order to determine whether or not to identify an object in a hazardous scenario. In such a scenario, the detection procedure should be run in OFF mode, meaning there should be no train at the platform.

\begin{figure}[h!]
\centering
\includegraphics[width=0.4\textwidth]{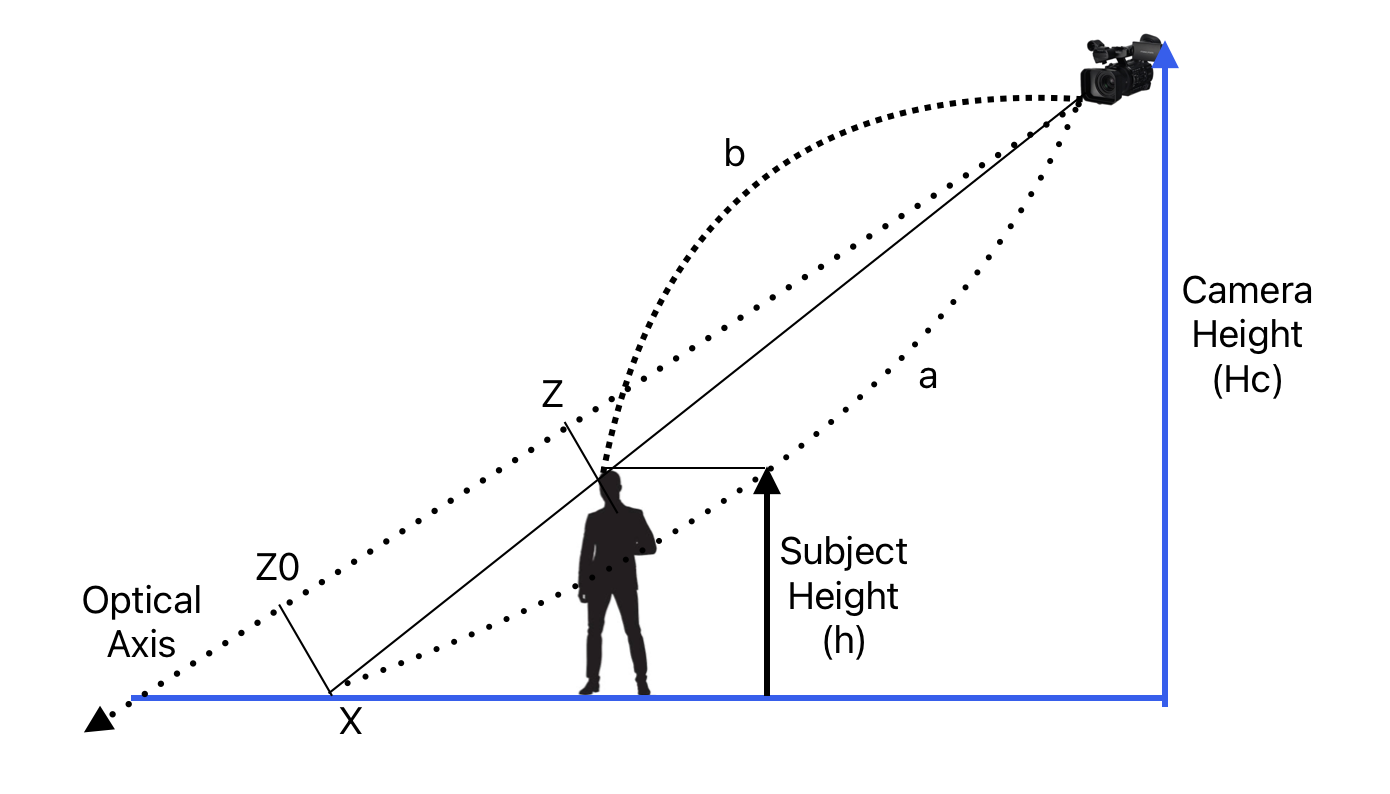}
\caption{Object Height }
\label{fig:obh}
\end{figure}

Finding the precise train states in the area for every single image is crucial to deciding on risky signs for object detection in the monitoring area. Clear definitions should be made of the train and monitoring regions. Monitoring area for passenger safety and risk region for train state monitoring. Generally speaking, a train at the station has four states, as table \ref{tab1} illustrates. I.e., state of approaching (IN-State), state of stopping (ON-State), state of pulling out (OUT-State), and state of being empty (OFF-State). Figure \ref{fig:obh} shows the concept of finding out the height of an object.

\begin{table}[]
\caption{A Train States At The Station}
\centering

\begin{tabular}{l|l}

\hline
Train States & \multicolumn{1}{c}{Description} \\ \hline
OFF          & There is no train               \\
IN           & Train is Approaching            \\
ON           & Train is arrived and stopped    \\
OUT          & Train is pulling out            \\ \hline
\end{tabular}
\label{tab1}
\end{table}

Position crossing the ground on the extension running from a camera to the subject's head is known as "$x$". This time, the ratio of distance "$a$" from the camera to the ground $X$ with respect to the distance "$b$" from the camera to the head is the same as the ratio of distance "$z0$" on the optical axis with respect to the distance "$z$". The following equation can be used to determine the subject in the case's height "$h$".

\begin{equation}
     h = H_c \times (1- \frac{b}{a}) = H_c \times (1- \frac{z}{z_0})
\end{equation}

By using this algorithm, we can determine the correct location for a safe distance for passengers so that they are not dangerous when they are on the platform while waiting for the train.

\section{EXPERIMENTS AND RESULT}
\begin{figure}[ht!]
\centering
\includegraphics[width=0.4\textwidth]{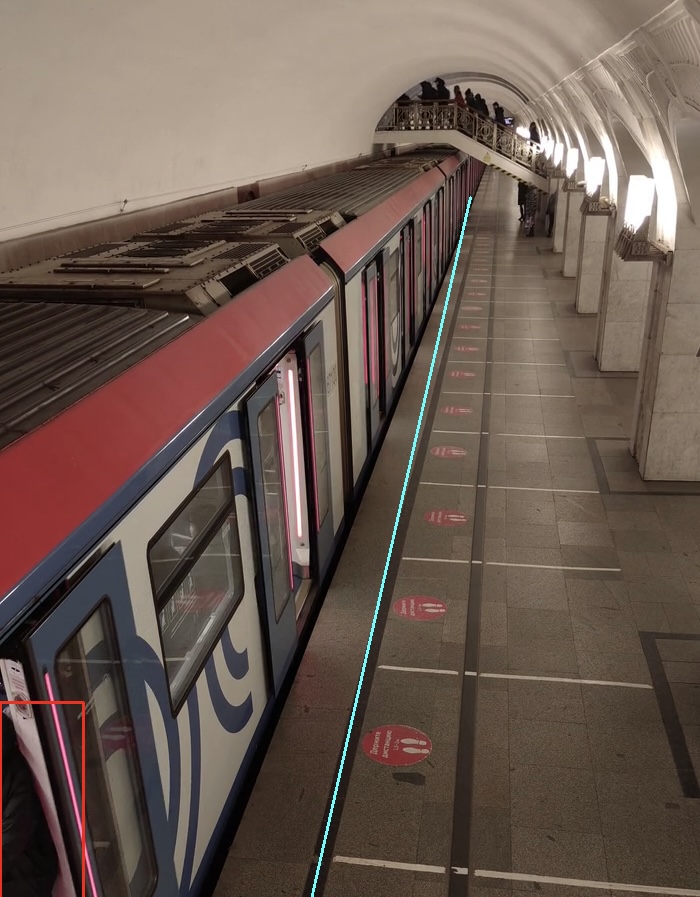}
\caption{Experiment Result: No Passenger}
\label{fig:ex0}
\end{figure}

To verify system performance, the experiment has been executed with an example video train situation at the Leninskiy Prospekt Metro Station. The experimental results are shown in the image \ref{fig:ex0} a result when there are no passengers, so the results obtained are that the system will not detect a single object except a passenger. 
Next Fig. \ref{fig:ex1} and   \ref{fig:ex2} show the results of object detection in the danger area. 

Fig. \ref{fig:ex1} and \ref{fig:ex2} show that there are several red boxes, which means that several people are crossing the yellow passenger line. so that it will provide a warning at the train station if an unauthorized person is detected, namely when someone crosses the yellow line without permission. As the popularity of railway stations increases and more and more people start relying on metro trains, automatic detection of unauthorized entries must be detected in real time, and appropriate action must be taken. We compared tests using 2 hardware, namely Hailo-8 and Jetson Orin Nano. The results are shown in Table \ref{tab2}. From this table, it can be concluded that the Hailo-8 AI accelerator had better Efficiency and lower latency than the Jetson Orin Nano. So by using the Hailo-8 AI accelerator, we can quickly detect passengers in dangerous situations at train stations.

\begin{table}[h!]
\caption{Result}
\begin{center}
\begin{tabular}{c c c c c c}
\hline
No. & Acc (\%) & Latency (ms) & Power (W) & Efficiency & Hardware \\
\hline
1 & 61.661 & 54.174 & 9.1 & 0.125 & Jetson Orin Nano \\
2 & 70.791 & 20.878 & 10.737 & 0.316 & Hailo-8 \\
\hline
\end{tabular}
\label{tab2}
\end{center}
\end{table}

\begin{figure}[ht!] 
\centering
\includegraphics[width=0.4\textwidth]{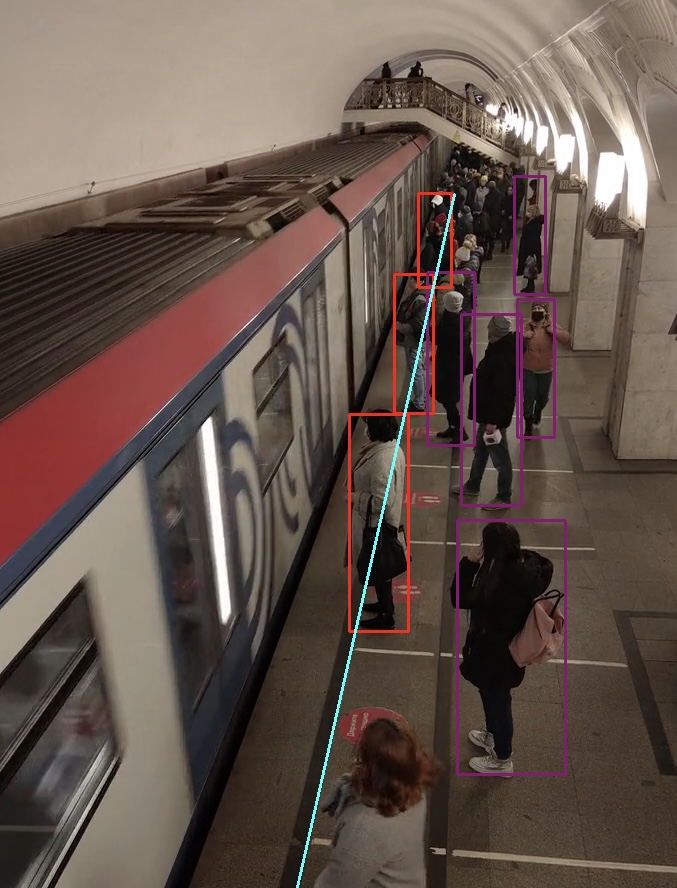}
\caption{Experiment Result : Passenger1}
\label{fig:ex1}
\end{figure}

\begin{figure}[h!]
\centering
\includegraphics[width=0.4\textwidth]{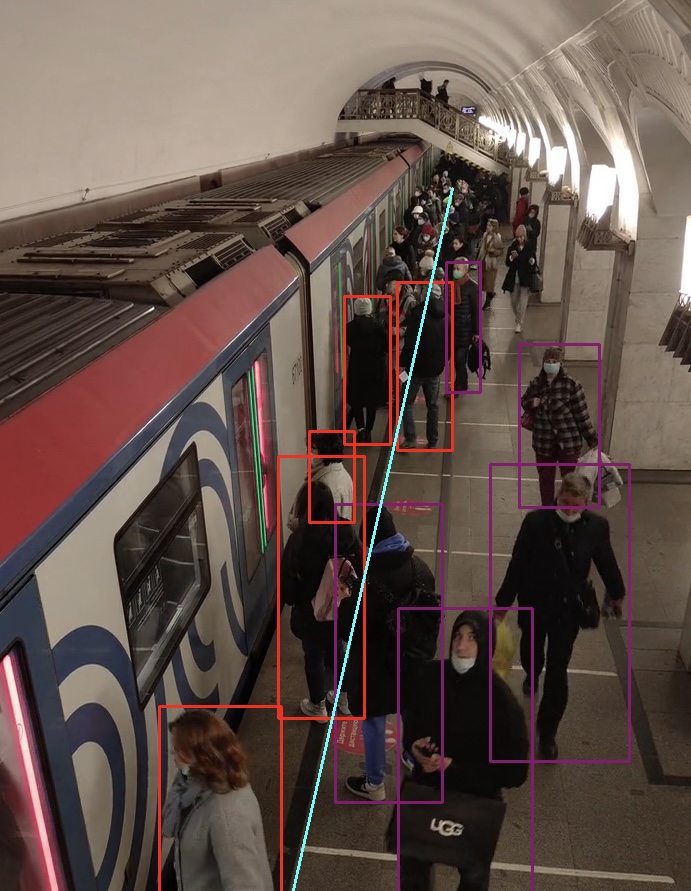}
\caption{Experiment Result : Passenger2}
\label{fig:ex2}
\end{figure}

\section{conclusion}
This paper aims to detect people who use YOLOX with AI Accelerator for train station security. The algorithm used is YOLOX and testing  by simulating conditions at the Leninskiy Prospekt Metro Train Station

The experimental results show that the YOLOX algorithm can detect train conditions and passenger detection and successfully provide a danger signal to passengers crossing the yellow line. From Table 2 it can be concluded that the simulation using the Hailo-8 AI accelerator has better efficiency and lower latency than the Jetson Orin Nano. So by using the Hailo-8 AI accelerator, we can Fast detect passengers in dangerous situations at train stations.

\section{ACKNOWLEDGEMENT}
This research was supported by National Science and Technology Council, Taiwan with Grant Number NSTC- 113-2221- E-305-018-MY3. The authors are grateful for supported in this research.

\vspace{12pt}
\color{black}

\end{document}